\documentclass[fleqn,11pt]{llncs}

\usepackage[authoryear,longnamesfirst]{natbib}

\RequirePackage{graphicx}
\RequirePackage{amsmath,amsfonts,amssymb}

\RequirePackage{expl3,xparse}
\RequirePackage{etoolbox,balance}
\RequirePackage{booktabs,makecell,multirow,array,colortbl,dcolumn,stfloats}
\RequirePackage{xspace,xstring,footmisc}
\RequirePackage[svgnames,dvipsnames]{xcolor}

\RequirePackage[colorlinks]{hyperref}

\usepackage{numprint}
\usepackage{listings}
\usepackage{subcaption}
\usepackage[utf8]{inputenc} 

\newcolumntype{R}[1]{>{\raggedleft\let\newline\\\arraybackslash\hspace{0pt}}m{#1}}
\newcolumntype{L}[1]{>{\raggedright\let\newline\\\arraybackslash\hspace{7pt}}m{#1}}

\newcounter{nalg}[section]
\renewcommand{\thenalg}{\arabic{nalg}}
\DeclareCaptionLabelFormat{algocaption}{\textbf{Algorithm \thenalg}}

\lstnewenvironment{algorithm}[1][]
{   
	\refstepcounter{nalg} 
	\captionsetup{labelformat=algocaption,labelsep=colon} 
	\lstset{ 
		mathescape=true,
		frame=tB,
		numbers=left, 
		numberstyle=\tiny,
		breaklines=true,
		postbreak=\mbox{\textcolor{red}{$\hookrightarrow$}\space},
		basicstyle=\ttfamily\footnotesize, 
		keywordstyle=\color{black}\bfseries,
		keywords={,input, output, return, datatype, function, if, else, foreach, while, begin, end, }
		numbers=left,
		xleftmargin=.03\textwidth,
		tabsize=2,
		#1 
	}
}
{}

\title{Hierarchy exploitation to detect missing annotations on hierarchical multi-label classification}

\author{
	Miguel Romero\inst{1} \and
	Felipe Kenji Nakano\inst{2,3} \and
	Jorge Finke\inst{1} \and
	Camilo Rocha\inst{1} \and
	Celine Vens\inst{2,3}
}

\institute{
	Department of Electronics and Computer Science,
	Pontificia Universidad Javeriana, Cali, Colombia
	\and
	Department of Public Health and Primary Care, KU Leuven Campus KULAK,		 
	Kortrijk, Belgium
	\and
	Itec, imec research group at KU Leuven,
	Kortrijk, Belgium
	\email{miguelangel.romero@javerianacali.edu.co}
}

\begin{document}

\maketitle
	
\begin{abstract}
	The availability of genomic data has grown exponentially in the last
	decade, mainly due to the development of new sequencing
	technologies. Based on the interactions between genes (and gene
	products) extracted from the increasing genomic data, numerous
	studies have focused on the identification of associations between
	genes and functions. While these studies have shown great promise,
	the problem of annotating genes with functions remains an open
	challenge. In this work, we present a method to detect missing
	annotations in hierarchical multi-label classification datasets. We
	propose a method that exploits the class hierarchy by computing
	aggregated probabilities to the paths of classes from the leaves to
	the root for each instance. The proposed method is presented in the
	context of predicting missing gene function annotations, where these
	aggregated probabilities are further used to select a set of
	annotations to be verified through \textit{in vivo} experiments. The
	experiments on \textit{Oriza sativa Japonica}, a variety of rice,
	showcase that incorporating the hierarchy of classes into the method
	often improves the predictive performance and our proposed method
	yields superior results when compared to competitor methods from the
	literature.
\end{abstract}

Genomic data has become exponentially available in the last decade,
mainly due to the development of new technologies, including gene
expression profiling generated with RNA
sequencing~\citep{ranganathan-bioinformatics-2019}. Based on the
interactions between genes (and gene products) extracted from the
increasing genomic data, numerous studies have focused on the
identification of associations between genes and
functions~\citep{rust-annoteq-2002,vandepoele-arabidopsis-2009,vandam-disease-2017}.

Biologically, this task is usually addressed through \textit{in vivo}
experiments that require a significant investment of money, time, and
effort given the combinatorial nature of the problem (i.e., a gene or
a gene product may be associated with multiple functions and the
number of functions is large~\citep{zhou-insilicoannot-2005}).
Approaches have emerged as alternatives to overcome these limitations
by combining mathematical, probabilistic, and computational methods
with biological data (see,
e.g.,~\cite{deng-prot-2003,luo-matrix-2007,jiang-hbn-2008,cho-functanalysis-2016,nakano-pannot-2019}). 
While these studies have shown great promise, the problem of
annotating genes with functions remains an open challenge.

Gene functions, defined by Gene Ontology (GO), are structured in a
hierarchy~\citep{go-go-2019}. For this reason, the problem of
predicting gene functions (or functional annotation) is generally
modeled as a hierarchical multi-label classification (HMC) task. The
objective of HMC is to fit a predictor that maps a set of instances
(e.g., genes of an organism) to a set of hierarchically organized
labels (e.g., biological processes in the GO hierarchy), while
respecting the hierarchy constraint among them. The \emph{hierarchy
	constraint} states that if an instance is predicted to have a
particular class, the ancestor classes must also be included in the
prediction of such an
instance~\citep{vens-hmc-2008,silla-hierarchy-2011}.

Gene functional annotation is generally addressed as a (hierarchical)
multi-label classification problem, under the assumption that the
functional information available (present annotations as well as the
absent ones) can be trusted and can be used as a training set to
construct an inductive model. However, it is well known that
functional information of genes (and genes products) is
incomplete~\citep{valentini-tpr-2009}. Thus, it is important to use
methods able to handle incomplete labeled data and that focus on
detecting annotations that are
missing~\citep{yu-weak-2015,sabzevari-noise-2018}.

This work introduces \textit{hieRarchical multi-labEl clAsSification
	to diScover mIssinG aNnotations} (\textit{REASSIGN}), a method to
detect missing annotations based on HMC that exploits the class
hierarchy to select a set of annotations (e.g., gene-function
associations). Its specific purpose is twofold: first, it can be used
to complete a given annotation dataset; second, the completed dataset
can be used to create better supervised models.

To this aim, HMC classifiers based on tree ensembles are used to
compute the probability of association of every instance-class pair
(e.g., genes and biological functions). For each gene, aggregated
probabilities are computed for the paths in the class hierarchy, where
a path is a sequence of classes with ancestral relations from a leaf
to the root of the hierarchy. Aggregated probabilities are used to
group paths of classes instead of using single instance-pair
associations. Based on the aggregated probabilities of the paths and
genes, a set of annotations absent in the given annotation dataset and
complying with the hierarchy constraint is selected as output.

The proposed method is evaluated on \textit{Oryza sativa Japonica}, a
variety of rice, and it is compared to different methods from the
literature~\citep{sabzevari-noise-2018,nakano-pannot-2019}. In
addition, eight HMC datasets of biological processes from the GO
hierarchy for rice are introduced. These datasets correspond to
subsets of rice genes and biological processes (GO sub-hierarchies),
whose features are structural properties and embeddings of the gene
co-expression network. The results show that the proposed method
outperforms the comparison methods in most cases. We find that
exploiting the hierarchy of functions helps to better identify
gene-function associations. The evidence suggests that this is a
promising approach for reducing the cost, time, and effort required
for experimental verification in a lab.

The remainder of the paper is organized as follows.
Section~\ref{sec:prelim} provides some theoretical preliminaries.
Section~\ref{sec:related} reviews related work.
Section~\ref{sec:method} introduces the method to detect missing
annotations exploiting the class hierarchy. Section~\ref{sec:setup}
describes the datasets and experimental setup for the gene function
prediction in \textit{Oryza sativa Japonica}, followed by the results
and discussion in Section~\ref{sec:results}. Finally,
Section~\ref{sec:concl} draws concluding remarks and future research
directions.

\section{Preliminaries}
\label{sec:prelim}

This section presents preliminaries on hierarchical multi-label
classification.

\subsection{Hierarchical Multi-label Classification}

Classification problems may be defined using binary, multi-class, or
multi-label prediction tasks, where predictions consist of a single
class, a single class from a set of mutually exclusive classes, and a
subset of classes, respectively.

Hierarchical multi-label classification is an extension of the
multi-label classification, that addresses the task of structured
output prediction where the classes are organized into a hierarchy
(such as the GO) and instances may belong to multiple
classes~\citep{vens-hmc-2008}. Formally, a HMC task is defined as
follows:

\begin{definition}
	Let $I$ be an instance space (set of instances) and $(C, \leq_h)$ a
	class hierarchy (where $C$ is a set of classes and $\leq_h$ is a
	partial order). The objective is to find a function
	$\psi:I\rightarrow2^C$ such that $c\in \psi(x)\implies\forall
	c'\leq_h c:c'\in \psi(x)$ (i.e., $\psi$ complies with the hierarchy
	constraint).
\end{definition}

~\cite{silla-hierarchy-2011} exposed that there are two types of
methods to explore the hierarchical structure. First, local or
top-down classifiers refer to partially predicting the classes in the
hierarchy from top to bottom taking into account the predictions of
parent classes. Second, global classifiers refer to a single
classifier that considers the entire hierarchy at once.

Overlooking the class relationships often leads to situations where
an instance is predicted to have a particular class, but the ancestor
classes are not included in the prediction. In other words, the
prediction does not satisfy the hierarchy constraint. Satisfying
ancestral constraints is also referred as the true-path rule in
biology~\citep{valentini-tpr-2009,ashburner-go-2000}.

\section{Related Work}
\label{sec:related}

In this section, we present a literature overview of studies on
hierarchical multi-label classification and prediction of gene (or
gene products) functions.

Some studies have focused on hierarchical multi-label classification
across different domains. For example,~\cite{dimitrovski-hmlc-2010}
presented a global approach that addresses HMC using random forests
of predictive clustering trees (PCTs) to annotate
images.~\cite{ramirez-corona-hmlc-2016} introduced a local approach
based on chained path evaluation. It used a classifier to train
non-leaf classes (i.e., classes with at least one descendant) in the
hierarchy, including information on ancestral relations through extra
features with the prediction of parent classes. 

Other studies have focus on the gene (or gene products) function
prediction problem. For example,~\cite{jiang-hbn-2008} proposed
Hierarchical Binomial-Neighborhood, a probabilistic and local HMC
approach to predict protein functions in yeast. Their results showed
that their method outperforms approaches based on independent class
prediction. However, it requires a high computational cost to compute
probabilities of every protein-function pair.~\cite{yu-weak-2015}
presented an approach to replenish the missing function labels and to
predict functions for unlabeled proteins in a hierarchical manner
assuming that the labeled data was incomplete. Their method combines
the hierarchical structure of functions and the similarity between
labels to identify interaction between proteins and functions using
guilt by association (see~\cite{petsko-gba-2009}).

~\cite{zhao-hphash-2019} presented Gene Ontology Hierarchy Preserving
Hashing, a gene function prediction method that retains the
hierarchical order between GO functions. It used a hashing technique
based on the taxonomic similarity between functions to capture the GO
hierarchy and predict gene functions. Their results showed that their
method preserved the GO hierarchy and helped to improve prediction
performance.~\cite{nakano-pannot-2019} performed a comparison among
publicly available HMC methods. According to their results,
Clus-Ensemble, a random forest of predictive clustering trees adapted
to HMC~\citep{schietgat-pct-2010}, provided superior results.
However, the authors did not explicitly propose a method to identify
missing annotations on HMC problems.

~\cite{zhou-gcn-2020} presented an approach to predict functions of
maize proteins, called Deep Graph Convolutional network model. It
used amino acid sequences of proteins and the GO hierarchy to predict
functions of proteins. Their results showed that their approach is a
powerful tool to integrate amino acid data and the GO structure to
accurately annotate proteins. Similarly,~\cite{cruz-enrich-2020}
aimed to predict the phenotypes and functions associated to maize
genes using (i) hierarchical clustering based on datasets of
transcriptome (set of molecules produced in transcription) and
metabolome (set of metabolites found within an organism), and (ii) GO
enrichment analyses. Their results showed that profiling individual
plants is a promising experimental design for narrowing down the
lab-field gap.

~\cite{romero-clust-2022} presented a method that combines the
functional information with the gene co-expression network of an
organism to extract features that capture the details of the GO
hierarchy using spectral clustering. Their results showed that the
extracted features are key to improve the performance of the gene
function prediction task on rice, using a global HMC approach of
random forests of decision trees.

Other studies addressed gene function prediction, obtaining
state-of-the-art performance for different case studies. However they
do not take into account hierarchical dependencies between classes
as they focus on multi-class problems instead (see,
e.g.,~\cite{abu-ngcn-2019,hamilton-induct-2017,kipf-gcn-2017,makrodimitris-afp-2020,chen-gfp-2021,xiao-gnn-2021}). 
Therefore, such studies can not be compared directly to assess
hierarchical multi-label classification.

\section{The proposed method}
\label{sec:method}

In this section, we introduce a definition of the problem of
predicting missing gene functions annotations and present a general
method to detect missing annotations in HMC problems.

\subsection{Problem definition}

Given a set of genes $V$, a set of biological functions $A$, and an
annotation function $\phi: V \rightarrow2^A$, where each gene is
associated with the collection of biological functions to which it is
known to be related (e.g., verified through \textit{in vivo}
experiments). The goal is to use the information represented by
$\phi$, together with additional information about $V$ (e.g., genomic
sequences or gene co-expression data), to obtain a function $\psi: V
\to 2^A$ that augments $\phi$ with previously undetected annotations.
The problem of predicting gene functions is generally addressed as a
HMC task, i.e., $V$ corresponds to the instance space $I$, the
biological functions are structured in a hierarchy $(A, \leq_h)$
(e.g., Gene Ontology hierarchy), and the functions $\phi$ and $\psi$
comply with the hierarchy constraint. Associations between genes
and functions not present in $\phi$ have either not (yet) been found
through \textit{in vivo} experiments or do not exist in a biological
sense. The new associations identified by $\psi$ are a suggestion of
functions that need to be verified through \textit{in vivo}
experiments. The function $\psi$ can be built from a predictor of
gene functions, e.g., based on a supervised machine learning model.

Formally, the gene function prediction problem is defined as follows:
\begin{definition}\label{def.network.ancoexp}
	Let $V$ be a set of genes, $A$ a set of biological functions, and 
	$\phi:V\rightarrow2^A$ a function describing known annotations. The
	objective is to obtain a function $\psi:V\to2^{A}$ that augments
	$\phi$ and complies with the hierarchy constraint.
\end{definition}

\subsection{REASSIGN}

Given a HMC problem with instances $I$ and a class hierarchy
$(C,\leq_h)$, we introduce \textit{hieRarchical multi-labEl
	clAsSification to diScover mIssinG aNnotations} (\textit{REASSIGN}),
a method to detect missing annotations for $I$.

The input of the method are a dataset $X$ comprising $|I|$ instances
and $|F|$ features, the class hierarchy represented as a tree with
$|C|$ vertices (e.g., biological functions of genes), and an
annotation function (e.g., $\phi$) represented as a label matrix $Y$
with an assignment of each instance in $I$ to a subset of classes
from $C$ (i.e., $Y:I\times C\rightarrow \{0,1\}$). The output of the
method is a suggestion of missing annotations in $Y$, i.e., a set of
annotations whose value in $Y$ is originally 0, but which are
believed to be false negatives. Naturally, the suggested annotations
must still satisfy the hierarchy constraint.

HMC datasets often have a large and imbalanced label set, specially
on deeper levels. In particular, despite being more informative,
deeper classes in the hierarchy have less annotations (are sparse),
leading to low predicted probabilities and predictive performance
overall. As a possible solution, we propose a method that exploits
the hierarchy of classes to compute an aggregated probability per
instance and path of classes in the hierarchy, relying on the
prediction probabilities provided by a HMC classifier. As a result,
at most $I\cdot p$ aggregated probabilities are computed
corresponding to all combinations between instances and paths, where
$p$ is the total number of paths from the leaves to the root in the
hierarchy. The aggregated probabilities of the paths are then used as
the criterion to select a set of annotations.

The proposed method consists mainly of 3 steps. First, a HMC
classifier is used to compute the probability of every instance-class
association, i.e., compute $Y':I\times C\rightarrow [0,1]$. Any local
or global HMC classifier can be used (e.g., tree ensembles or neural
networks), providing that the hierarchy constraint is satisfied.

Second, aggregated probabilities are computed for each instance and
each path from the leaves to the root in the class hierarchy by using
the predicted probabilities of the annotations in $Y'$. That is, only
the paths that go to leaves in the class hierarchy are considered,
because the addressed problem is leaf
mandatory~\citep{silla-hierarchy-2011}. Importantly, only new
potential annotations are used to compute aggregated probabilities,
i.e., instance-class associations satisfying \[(\forall i,c:i\in I
\land c\in C|Y [i,c]=0 \land Y'[i,c]>0).\] Different ways of
aggregating the probabilities can be used; in this work, we used the
average, sum, and minimum. Each aggregation function is considered an
independent variation of the proposed method:

\begin{itemize}
	\item \textbf{REASSIGN (min)}: aggregates probabilities by using the
	minimum probability along the path considered. Paths are identified
	by their most informative (deepest) class;
	\item \textbf{REASSIGN (sum)}: aggregates probabilities by using the
	sum of the probabilities along the path considered. The longer and 
	deeper the path, the larger the sum, and therefore, more informative 
	classes are considered;
	\item \textbf{REASSIGN (average)}: aggregates probabilities by using
	the average of the probabilities along the path considered. It
	balances the probabilities of the more informative (deeper) and less
	informative (shallower) classes.
\end{itemize}

\begin{figure*}[tbph!]
	\centering
	\includegraphics[width=\linewidth]{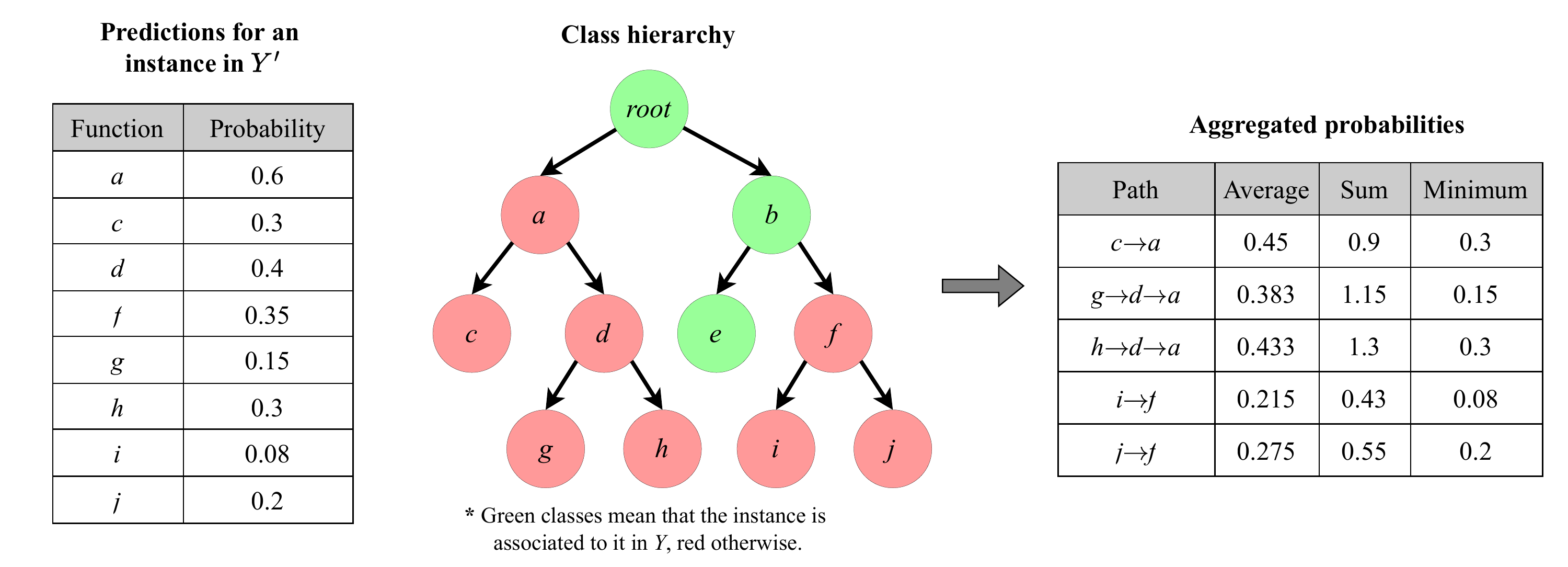}
	\caption{Given an instance, aggregated probabilities are computed
		for every path in the class hierarchy from the leaves to the root
		using the average, sum and minimum of the probabilities of the
		annotations in the path. Only those annotations whose value is 0 in
		$Y$ are considered in the paths. This Figure is best viewed in colors.}
	\label{fig:pathex}
\end{figure*}
These variations are illustrated in Figure~\ref{fig:pathex} where we
exemplify how aggregated probabilities of the paths in the hierarchy
are computed for a given instance. Classes coloured with green denote
that the instance $i\in I$ is already associated to the class $c\in
C$ in $Y$ (i.e., $Y[i,c]=1$), while classes coloured in red denote that
the instance is not associated to the class (i.e., $Y[i,c]=0$). Since
the method is focused in detecting missing annotations (i.e.,
instance-class pairs associations that have not been identified), only
classes coloured with red are used to compute aggregated probabilities.
Reinforcing the objective of detecting missing annotations through 
new paths in the hierarchy. Note that the path $d\rightarrow a$ is not 
considered since it does not go to a leaf node.

\begin{algorithm}[caption={Hierarchical multi-label classification to 
		discover missing annotations (REASSIGN)}, label={alg:hpe}, deletekeywords={function}]
input: 
	$X$: dataset
	$(C,\leq_h)$: class hierarchy 
	$Y:I\times C\rightarrow\{0,1\}$: instance-class associations
	$f(\_)$: aggregation function
	$N_p$: Number of paths to select
output: 
	$top\_annot\subseteq\{(i,c):i\in I\land c\in C\}$: subset of annotations 

compute $Y':I\times C\rightarrow[0,1]$ using a HMC method s.t. $Y'$ complies to the hierarchy constraint.
set $all\_paths=\{\}$
foreach instance $i\in I$
	foreach $path\in(C,\leq_h)$
		set $probs=\{\}$ and $annot=\{\}$
		foreach $c\in path$
			if $Y[i,c]=0\land Y'[i,c]>0$
				add $Y'[i,c]$ to $probs$
				add $(i,c)$ to $annot$
			end
		end
		add $(f(probs),annot)$ to $all\_paths$
	end
end
sort $all\_paths$ in decreasing order
set $top\_paths=all\_paths[0\dots N_p)$
set $top\_annot=\{\}$
foreach $(x, annot)\in top\_paths$
	add $annot$ to $top\_annot$ 
end
remove duplicates from $top\_annot$
return $top\_annot$ 
\end{algorithm}

At last, some paths are selected using their aggregated probability
as selection criterion. The number of paths to be selected is a
parameter denoted as $N_p$. All annotations within the top $N_p$
paths with higher aggregated probability are selected. Note that
there might be common annotations between the paths, so duplicates
have to be removed. For instance, paths $i\rightarrow f$ and
$h\rightarrow f$ in Figure~\ref{fig:pathex} share the class $f$,
hence in case both paths are selected, the association between the
instance and class $f$ will be duplicated. The resulting number of
annotations is denoted as $N$ and used for comparison with other
methods. A detailed description of the proposed method is presented
in Algorithm~\ref{alg:hpe}.

\section{Experimental setup}
\label{sec:setup}

In this section, a detailed description of the employed databases,
comparison methods, and evaluation measures is presented. 

\subsection{Datasets}

Two datasets are built using the functional information and the gene
co-expression network (GCN) for \textit{Oryza sativa Japonica}, a
variety of
rice~\citep{kurata-oryzabase-2006,childs-gcn-2011,sakai-rapdb-2013}.

The functional information depicts associations between genes and
functions previously identified through \textit{in vivo} or
\textit{in silico} experiments. For this work, the functional
information is imported from DAVID Bioinformatics
Resources~\citep{huang-david-2009}, that contains annotations of
biological processes, i.e., pathways to which a gene contributes.
Note that genes may be associated to several biological processes,
and biological processes may be associated to multiple genes. The
datasets are built using two different versions of the functional
information of rice. That is, each dataset has a different label
matrix, but they share the instances and features. One dataset (the
older version) is used to train and build the models, whereas the
other one is used to evaluate the detected missing annotations.

The first version of the functional information is from 2018 and it
comprises \numprint{3531} biological processes and \numprint{6367}
hierarchical relations that are part of the GO
hierarchy~\citep{go-go-2019}. A total of \numprint{197194}
associations between genes and functions are considered in this
version. The second version is from 2021 and since it is only used
for performance evaluation, the same set of biological processes and
hierarchical relations are considered. This version comprises a total
of \numprint{289407} associations.

The GCN is built using the co-expression information imported from
the ATTED-II database~\citep{obayashi-atted2018-2018} (version r17c).
A GCN is represented as an undirected and weighted graph where each vertex
represents a gene and each edge the level of co-expression between
two genes~\citep{aoki-gcn-2007,vandepoele-arabidopsis-2009}. The GCN
of rice $G = (V, E, w)$ comprises \numprint{19663} vertices (genes)
and \numprint{550813} edges. In this case, a mutual rank threshold of
$100$ is used as the cut-off measure for $G$, i.e., $E$ contains
edges $e$ that satisfy $w(e) \leq 100$. Note that the lowest value is
assigned to the strongest connections.

Biological processes are a subset of the functions in the GO
hierarchy, where each function in the topmost level represents a
sub-hierarchy. However, as functions can have more than one parent,
sub-hierarchies might not be independent (i.e., functions might
belong to multiple sub-hierarchies) and there might be several paths 
between two functions. The topological-sorting traversal algorithm
presented by~\cite{romero-hmc-2022} is used to transform the
hierarchy into a tree so that there is unique path between all pair
of function in the sub-hierarchies and all sub-hierarchies are
independent. Each sub-hierarchy is denoted as $H=(A,\leq_h)$, where
$A$ is the subset of biological processes and $\leq_h$ the binary
relation representing ancestral relations between pairs of
biological processes, i.e., $a\leq_hb$ means that function $b$ is
parent of function $a$ in the sub-hierarchy.

As a result, 8 sub-hierarchies of biological processes are used.
Table~\ref{tab:func} describes each sub-hierarchy $H$, starting by
the root term and its description, followed by the number of
biological processes $A$, the number of genes, the number of new
annotations (i.e., 0s that became 1s from 2018 to 2021 version), and
the number of functions per level. Note that the functional
information from 2021 includes more annotations (i.e., 0s that became
1s), but also drops some of them (i.e., 1s that became 0s).
Annotations are dropped from one version to other because it was
experimentally verified that such associations between genes and
functions do not exist. The prediction approach is applied to each
sub-hierarchy $H$ independently.

\begin{table*}[htbp!]
	\centering
	\resizebox{\textwidth}{!}{%
	\begin{tabular}{|p{1.8cm}|p{5cm}|r|r|r|p{5.3cm}|}
		\hline
		Root & Description & Functions & Genes & $0\rightarrow1$ & Functions per level\\
		\hline\hline
		GO:0032501 & multicellular organismal process & 26 & 538 & 184 & 8/10/6/1 \\  
		GO:0019752 & carboxylic acid metabolic process & 63 & 505 & 180 & 7/15/23/15/2 \\  
		GO:0032502 & developmental process & 68 & 871 & 537 & 10/19/25/11/2 \\  
		GO:0006796 & phosphate-containing compound metabolic process & 71 & 1142 & 669 & 9/16/22/16/7 \\  
		GO:0051179 & localization & 112 & 1285 & 1630 & 4/9/21/31/23/16/5/1/1 \\  
		GO:0065007 & biological regulation & 291 & 2137 & 2879 & 3/19/57/94/69/23/13/8/3/1 \\  
		GO:0008152 & metabolic process & 514 & 5348 & 14601 & 14/47/149/98/72/60/45/18/8/1/1 \\  
		GO:0009987 & cellular process & 594 & 5867 & 16520 & 32/67/117/144/93/66/46/16/10/1/1 \\ 
		\hline
	\end{tabular}
	}	
	\caption{Resulting sub-hierarchies of biological processes for rice.
		The identifier and description of each root function $r$ is
		presented in the first and second columns, respectively. The
		following columns show the number of functions $A$ within each
		sub-hierarchy, the number of genes associated to it, and the number
		of new annotations (i.e., 0s that became 1s). The last column shows
		the number of functions per level, e.g., the first sub-hierarchy has
		4 levels and there are 8, 10, 6, and 1 function on each level.}
	\label{tab:func}
\end{table*}

For each sub-hierarchy, we compute and combine two sets of features:
structural properties and node embeddings of the GCN. Given a
sub-hierarchy $H$ and its associated genes, structural properties of
the GCN are computed as features. In this case, the properties
included for each gene $u$ are the following:

\begin{itemize}
	\item degree: number of edges incident to $u$ (including $u$);
	\item average neighbor degree: average degree of the neighbors of $u$;
	\item eccentricity: maximum shortest distance from $u$ to any node
	in its connected component;
	\item clustering coefficient: ratio between the number of triangles
	(3-loops) and the maximum number of 3-loops that could that pass
	through $u$; 
	\item closeness centrality: reciprocal of the average shortest
	path length from $u$;
	\item betweenness centrality: the amount of influence that $u$ has
	over the interactions of other nodes in the network, measured as the
	number of shortest paths that pass through $u$;
	\item  Kleinberg's hub scores: defined as the principal eigenvector
	of $\mathbf{A}\mathbf{A}^T$, where $\mathbf{A}$ is the adjacency
	matrix of the graph~\citep{kleinberg-authority-1999}. Hubs are
	vertices linked to many other vertices.
	\item Kleinberg's authority score: defined as the principal
	eigenvector of $\mathbf{A}^T\mathbf{A}$. Authorities are the
	most central vertices on a network, which are connected to many
	different hubs.
	\item coreness: the highest order $k$-core containing the vertex $u$,
	where a $k$-core is a maximal subgraph in which each vertex has at
	least degree $k$.
\end{itemize}

These measures are computed using igraph~\citep{ju-igraph-2016}, an
open source and free collection of network analysis tools.
Additionally, a low-dimension embedding of the GCN is computed to
capture gene expression patterns. Embeddings are continuous
representations of nodes into a low-dimensional space that captures
node similarity and network structure. The goal is for properties in
the embedded representation to approximate properties in the original
network~\citep{grover-node2vec-2016}. In other words, embeddings are
vector representations that capture characteristics of the nodes by
using less data, thus being more tractable for machine learning. The
dimension of the embedding for each sub-hierarchy corresponds to the
number of biological processes in it (i.e., $|A|$).

\subsection{Comparison methods}

In this work, we employ 6 methods for comparison. More specifically,
we present a comparison method from the literature, followed by
2 baseline methods and 3 variants of our proposed method.

Despite providing insights on how prediction probabilities may be
used,~\cite{nakano-pannot-2019}, the most recent work in this
context, did not explicitly propose a method to identify missing
annotations on HMC problems. These authors have, however, showed that
random forests have superior predictive performance than other HMC
methods. For this reason, we used a global HMC classifier based on
random forests of decision trees as the baseline classifier for all
methods (including the proposed one), where all functions of the
sub-hierarchy are considered at once. The parameter values used for
random forest classifiers are: 200 estimators
(\textit{n\_estimators}) and minimum number of samples of 5
(\textit{min\_samples\_split}), whereas the number of folds used is
$k=5$.

The work of~\cite{yu-weak-2015} could be employed as a comparison,
nonetheless the authors addressed the problem of identify missing
functional annotations of proteins using a probabilistic model based
on similarities between functions and guilt by association. Their
method associates a protein and a function based on the correlation of
functions in the hierarchy and the information of related proteins in
the protein-protein interaction network (i.e., guilt by association).
Thus, it can not be seen or extended as a HMC method.

Apart from that, the literature presents several works that are
capable of identifying missing or wrong annotations in binary
classification~\citep{cao-nosie-2012,sluban-ensemble-2014,sabzevari-noise-2018,samami-noise-2020,zhang-miss-2020}. 
Unfortunately, these works require adaptation since they were
evaluated in the context of binary classification. Among these, the
recently proposed method presented by~\cite{sabzevari-noise-2018}
seems to be the the most related to the proposed method, since it
relies specifically on random forests, and it can be straightforwardly
adapted to HMC. In this work, this method is referred to as
\textit{Noise detect}.

Below, we present a more detailed description of each method
included in our experiments:

\begin{itemize}
	\item \textbf{Noise detect:}~\cite{sabzevari-noise-2018} recently
	proposed a method that employs an ensemble of decision trees to
	identify mislabeled instances in binary classification. More
	specifically, an instance is marked as noise if its
	misclassification rate is higher than a threshold, where the
	misclassification rate is defined as the proportion of predictors in
	which the instance is misclassified based on the number of
	predictors where the instance is out-of-bag. In this work, we adapt
	this method by selecting the top $N$ annotations with higher
	misclassification rate;
	
	\item \textbf{No aggr:} A variant of our proposed method that
	does not consider the hierarchy of classes. That is, no aggregation
	method is employed, and the predictions are employed directly from
	the classifier. This variant is included to highlight the importance
	of the hierarchical relationships;
	
	\item \textbf{Random}: A baseline random method that selects
	annotations without any criterion and complies with the hierarchy
	constraint. This method is included as reference point to validate
	the use of machine learning methods;
	
	\item \textbf{REASSIGN (min)}: A variant of our proposed method that
	aggregates probabilities by using the minimum probability along
	the path considered;
	
	\item \textbf{REASSIGN (sum)}: A variant of our proposed method that
	aggregates probabilities by using the sum of the probabilities
	along the path considered;
	
	\item \textbf{REASSIGN (average)}: A variant of our proposed method
	that aggregates probabilities by using the average of the
	probabilities along the path considered;
	
\end{itemize}

Since \textit{Noise detect} was built for binary classification
problems, a default threshold of 0.5 is used to define the labels.
However, in HMC, the predicted probabilities vary according to their
level in the hierarchy, the deeper a class is, the lowest the
probabilities get. For this reason, we adapt this method by using a
different threshold for each function according to its level in
the sub-hierarchy. The threshold is set as $t=0.5\cdot 0.75^{l-1}$
(similar to weights proposed by \cite{vens-hmc-2008}), where $l$ is
the level of the function. For instance, a threshold $t=0.5$ is used
for functions in the first level and $t=0.88989$ is used for
functions in the seventh level.

\subsection{Evaluation measures}

The performance evaluation of the methods is based on the true
positive and false positive measures, because the aim of this work is
to detect missing annotations (i.e., identifying 0s that became 1s)
and there are two versions of the datasets available that allows us
to verify the predictions. That is, the evaluation is focused on
annotations that are not detected, but might show up in the future
and can be verified using the newer version of the datasets.

We use precision@$N$ as the first evaluation metric. Given $N$, the
number of annotations selected in a dataset (derived from the number
of paths to be selected $N_p$), the precision for the selected
annotations is computed as %
\begin{equation*}
\textnormal{precision}@N=\frac{tp_N}{tp_N+fp_N}.
\end{equation*}

Moreover, different values of $N$ are used to avoid bias in the
prediction and evaluation, hence the number of selected paths can be
set according to the resources available for \textit{in vivo} biological
experimentation.

In addition, we use the area under the curve generated between the
different values of $N$ (in the $x$-axis) and the precision@$N$ (in
the $y$-axis) as second evaluation metric, denoted as, AUP@NC. This
metric aims to analyze the performance of the methods regardless of
the value of $N$ and to remove subjectiveness caused by using only one value
for each method and dataset. The area under the precision@$N$ curve
is defined as
\begin{equation*}
\label{equation:area1}    
\text{AUP@NC} = \sum_{N_i} \text{area}(N_i,\text{precision}@N_i)
\end{equation*}
where $N_i$ represents the different values of $N$.

The number of paths to be selected $N_p$ by the proposed method is
defined as a proportion of the number of 1s in the older version of
the dataset, i.e., the total number of associations between genes and
functions occurring in $Y$. In particular, we set $n\in[0,1]$ as the
proportion of occurring annotations and $N_p=\sum Y\cdot n$. The 
number of annotations within the top $N_p$ paths after
removing duplicates is denoted as $N$. That is, $N$ and $n$ have a positive
proportional relationship, i.e., lower values of $n$ lead to lower
values of $N$, and higher values of $n$ lead to higher values of $N$.
We use 20 different values of $n$, from $n=0.01$ incremented in steps
of 0.01 up to $n=0.2$. Note that the values of $N_p$ and $N$ are
different for each dataset, whereas the values of $n$ are the same
for all datasets.

Furthermore, in order to provide statistical evidence, the
Friedman-Nemenyi test is used. At first, the Friedman test verifies
if any of the compared methods performs statistically significantly
different from others. Next, the Nemenyi test ranks the methods where
methods with superior results are ranked in higher positions.
Graphically, methods connected by a horizontal bar, of length less or
equal to a critical distance, are not statistically significantly
different. As input to this test, we employ the area under the
precision@$N$ curve.

\section{Results and discussion}
\label{sec:results}

In this section, the experiments and results are presented. At first,
we analyze the predictive performance of the proposed and comparison
methods using the precision@$N$ measure, followed by a discussion on
how our method differs from its comparison counterparts through the
area under the precision@$N$ curve. Lastly, we analyze the
deepness of the annotations predicted by the proposed method.

\subsection{Comparison between all methods of the precision@$N$}

\begin{figure*}[tbph!]
	\centering
	\includegraphics[width=.4\linewidth]{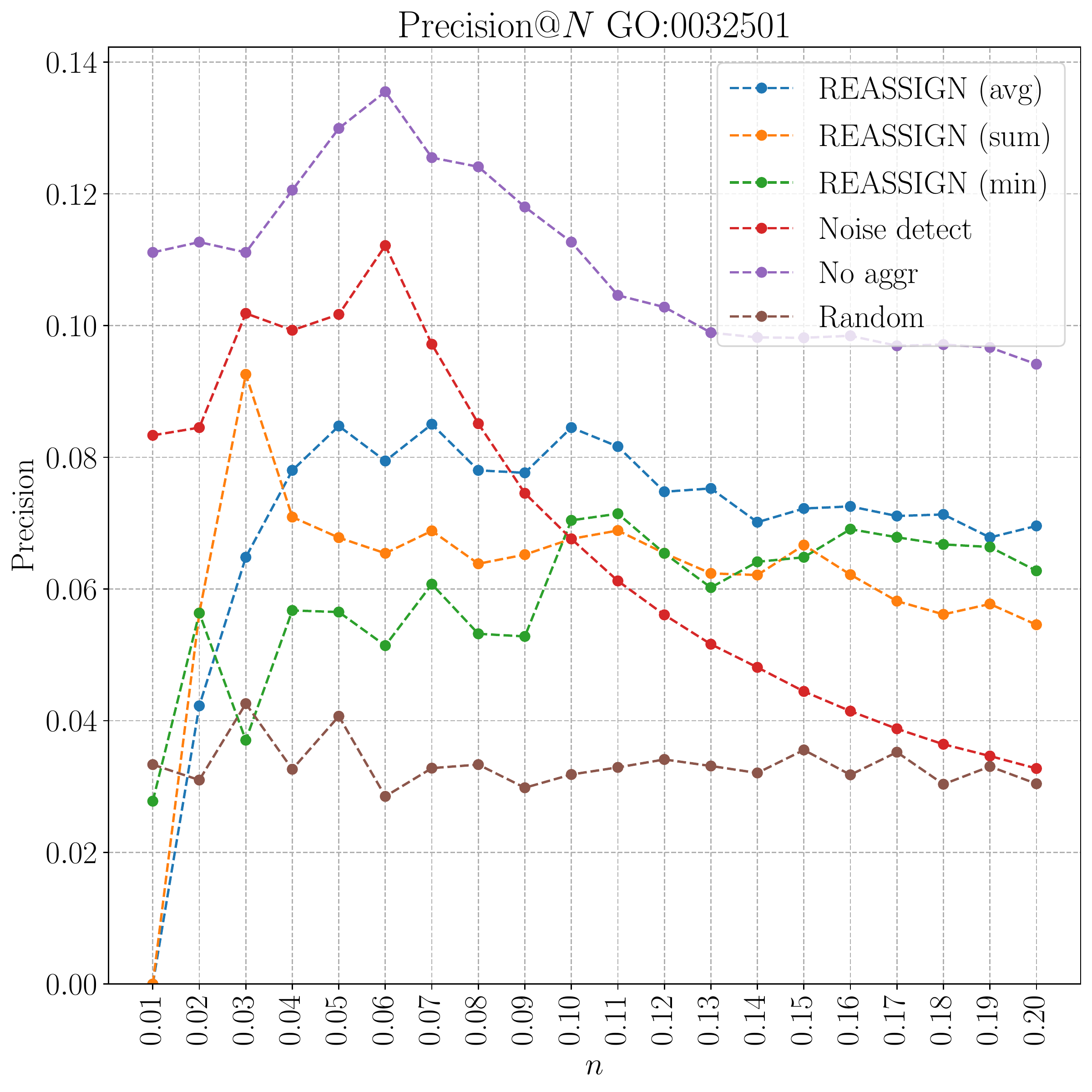}\hspace{1cm}
	\includegraphics[width=.4\linewidth]{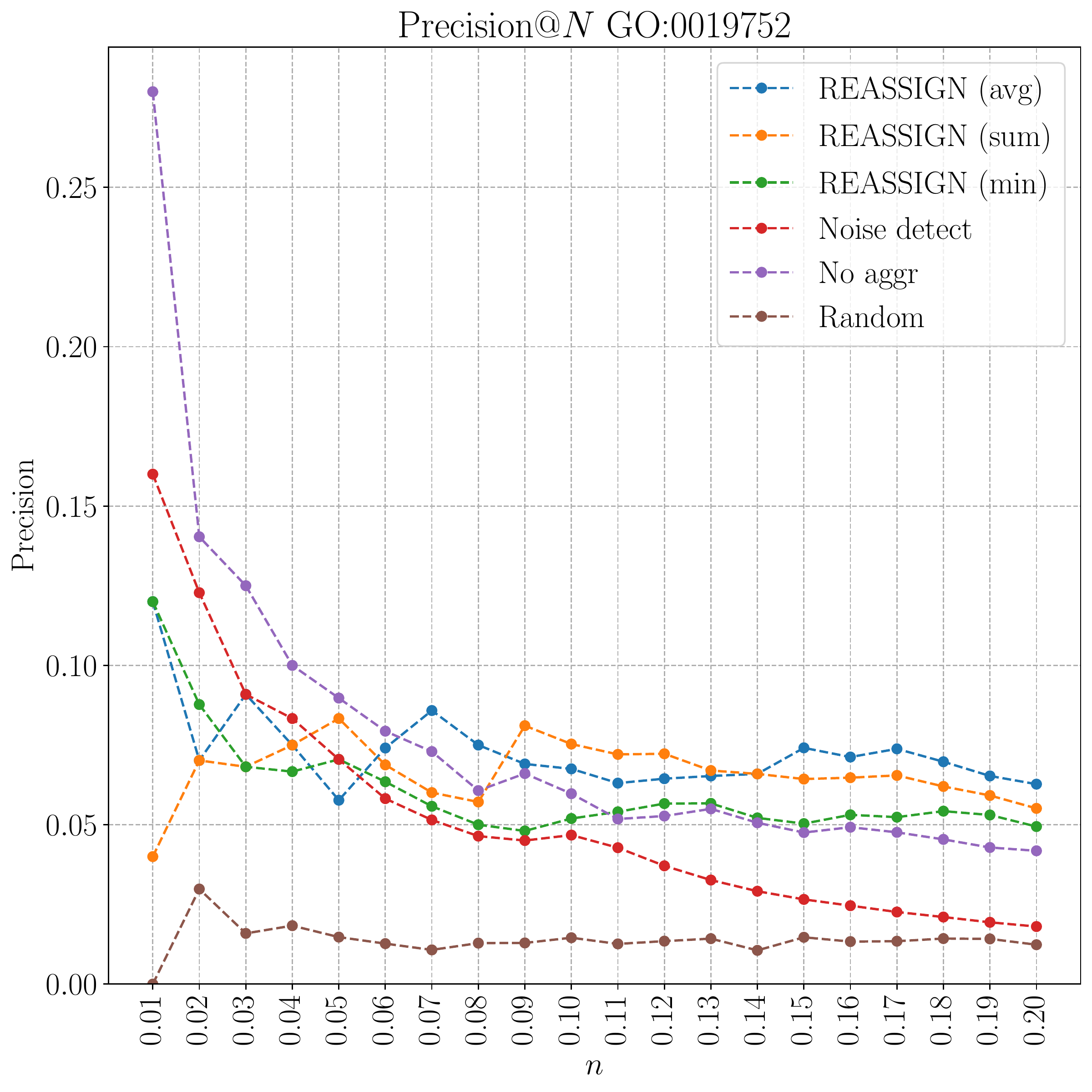}
	\includegraphics[width=.4\linewidth]{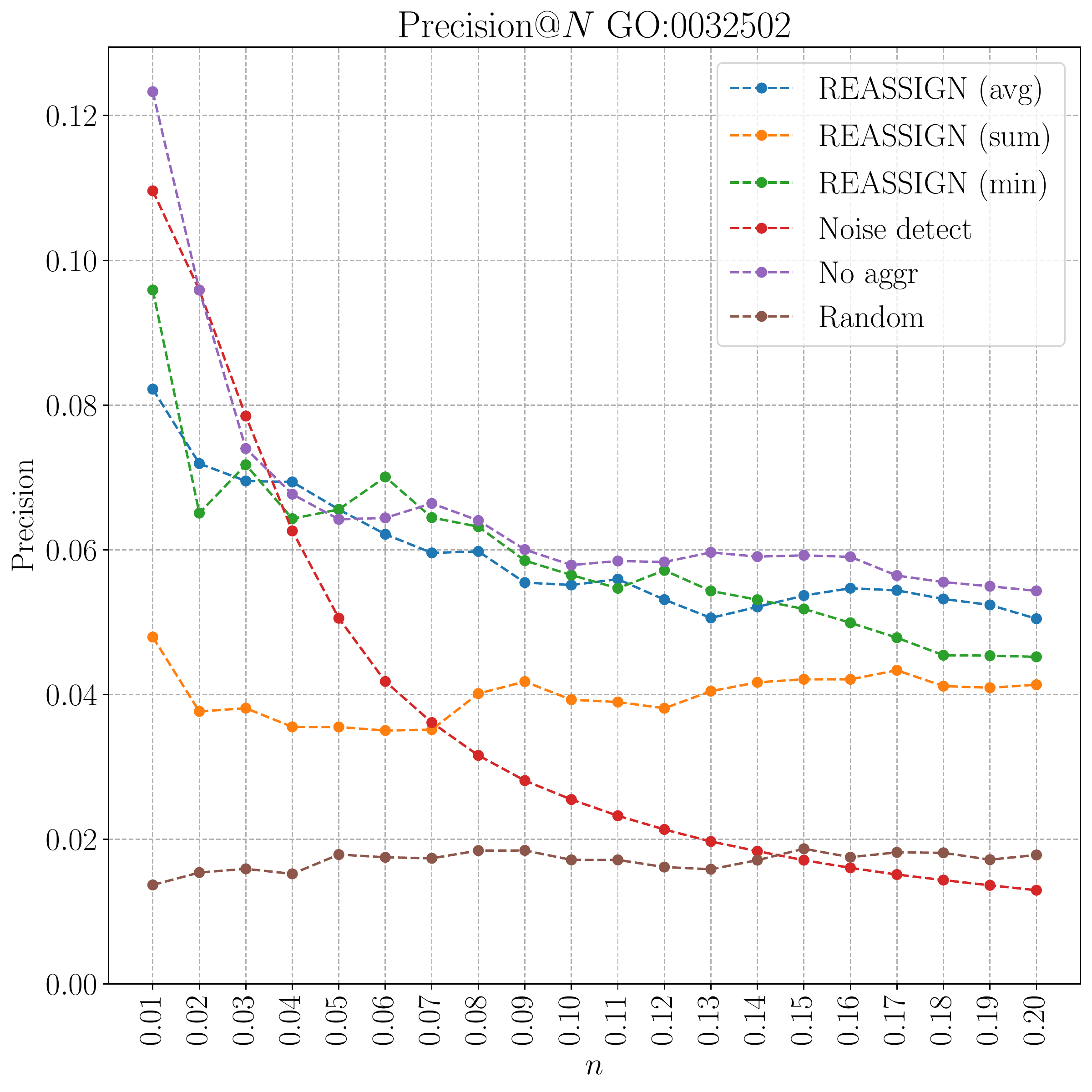}\hspace{1cm}
	\includegraphics[width=.4\linewidth]{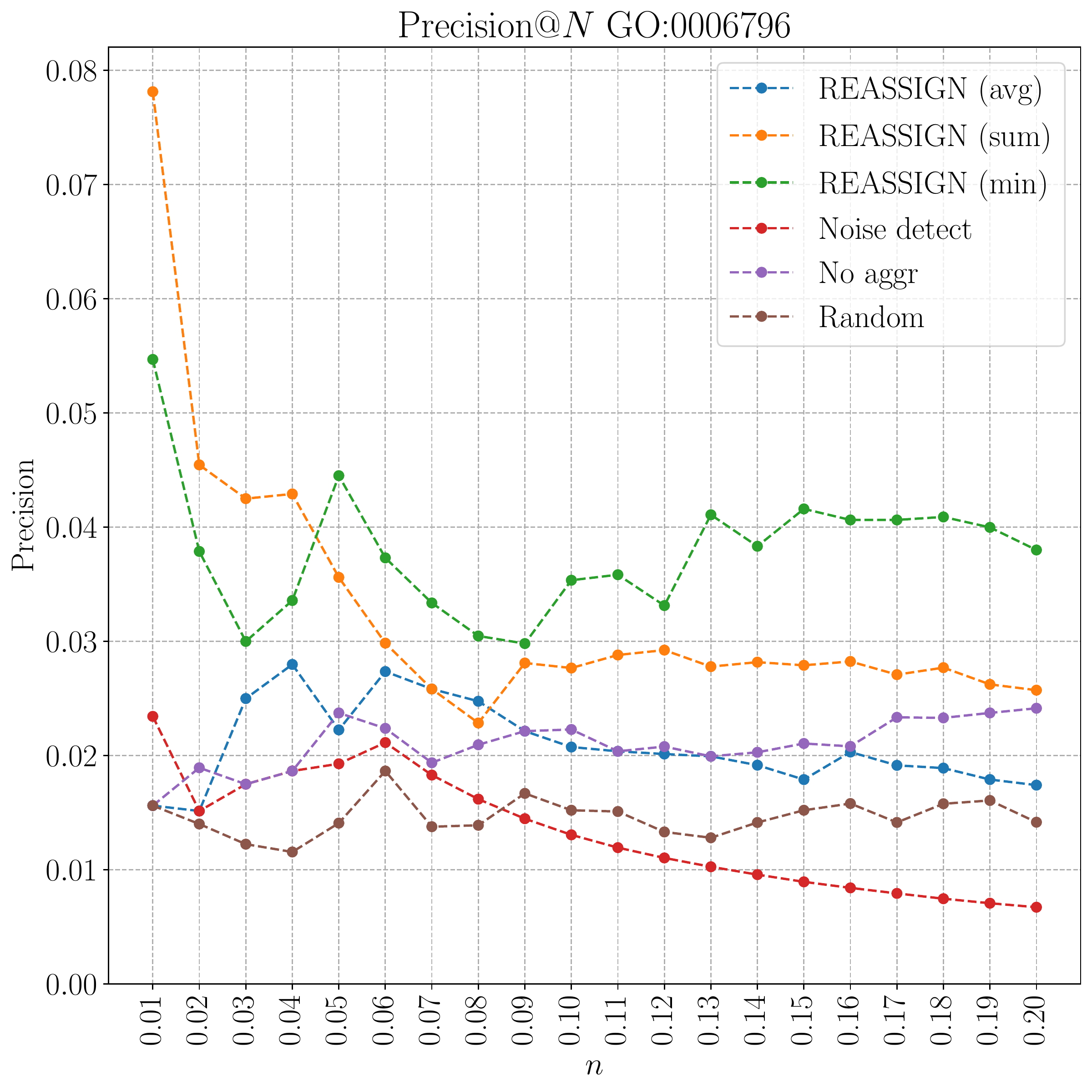}
	\includegraphics[width=.4\linewidth]{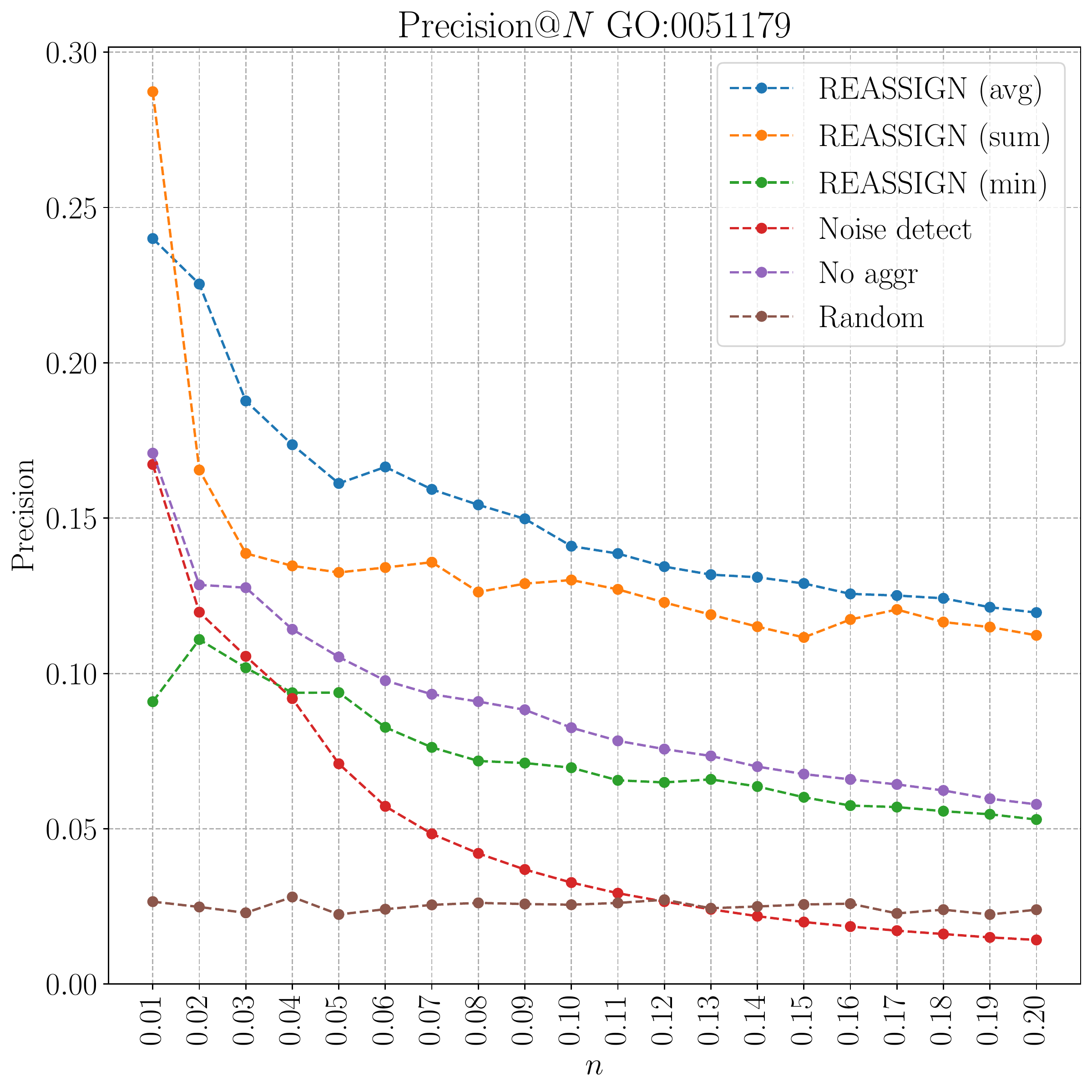}\hspace{1cm}
	\includegraphics[width=.4\linewidth]{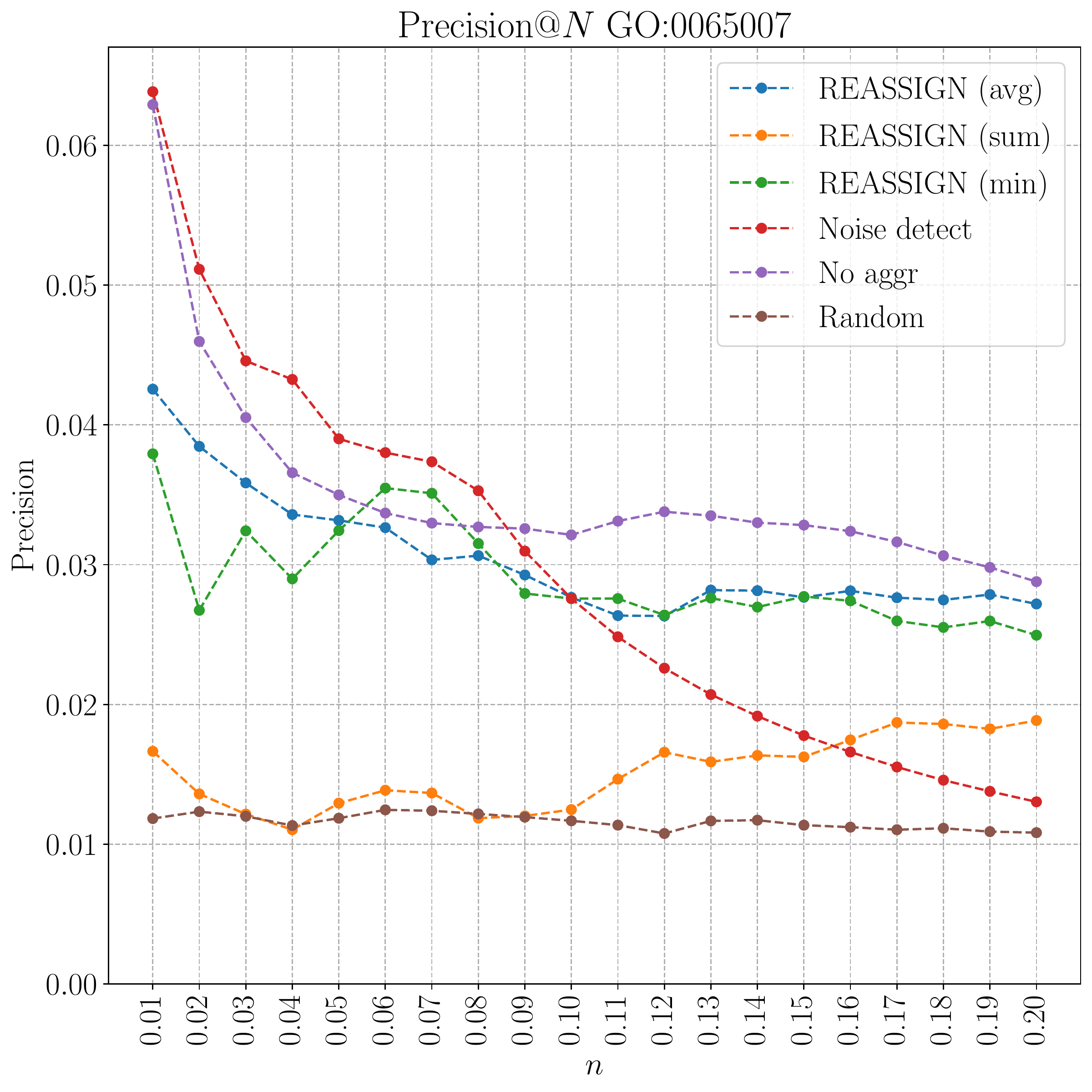}
	\includegraphics[width=.4\linewidth]{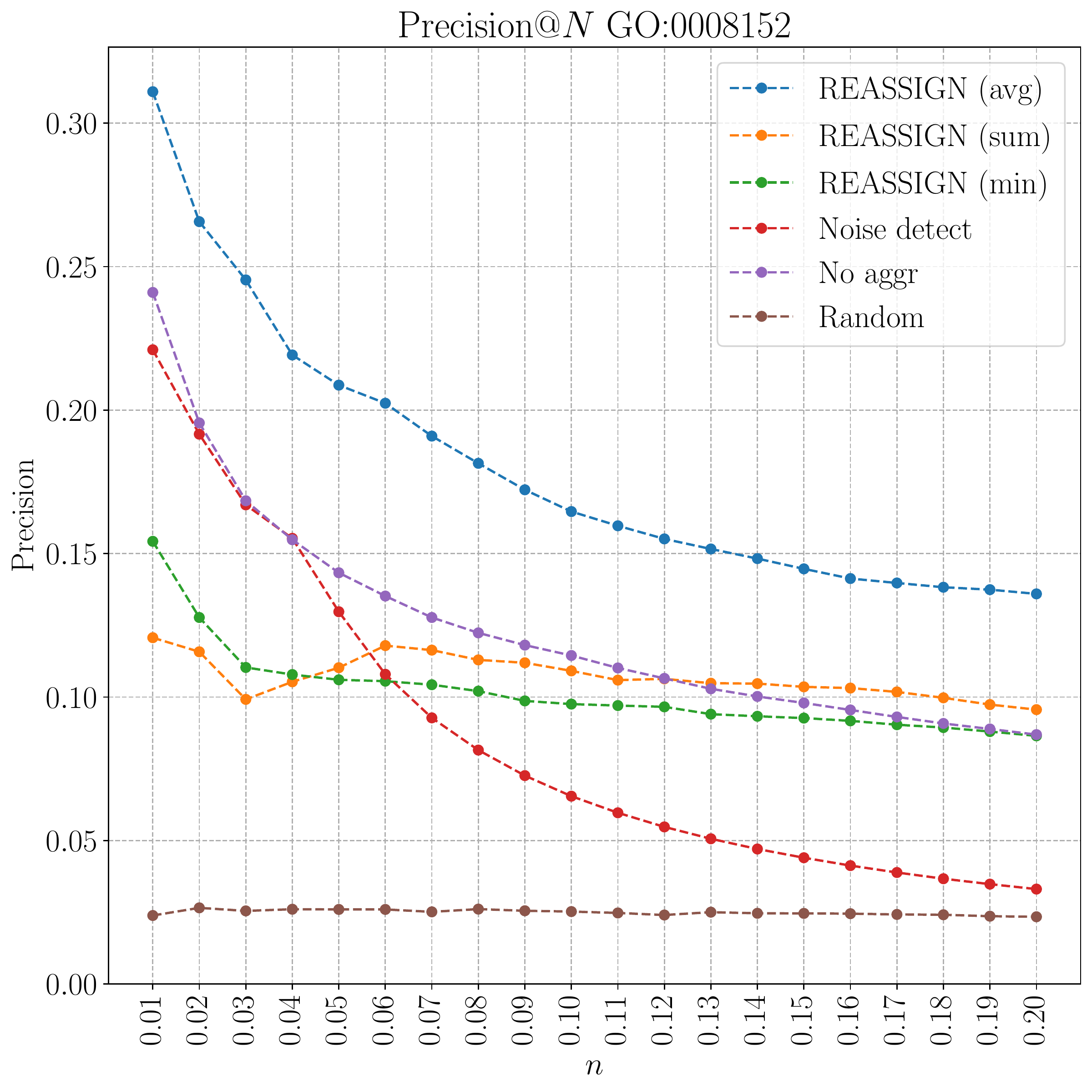}\hspace{1cm}
	\includegraphics[width=.4\linewidth]{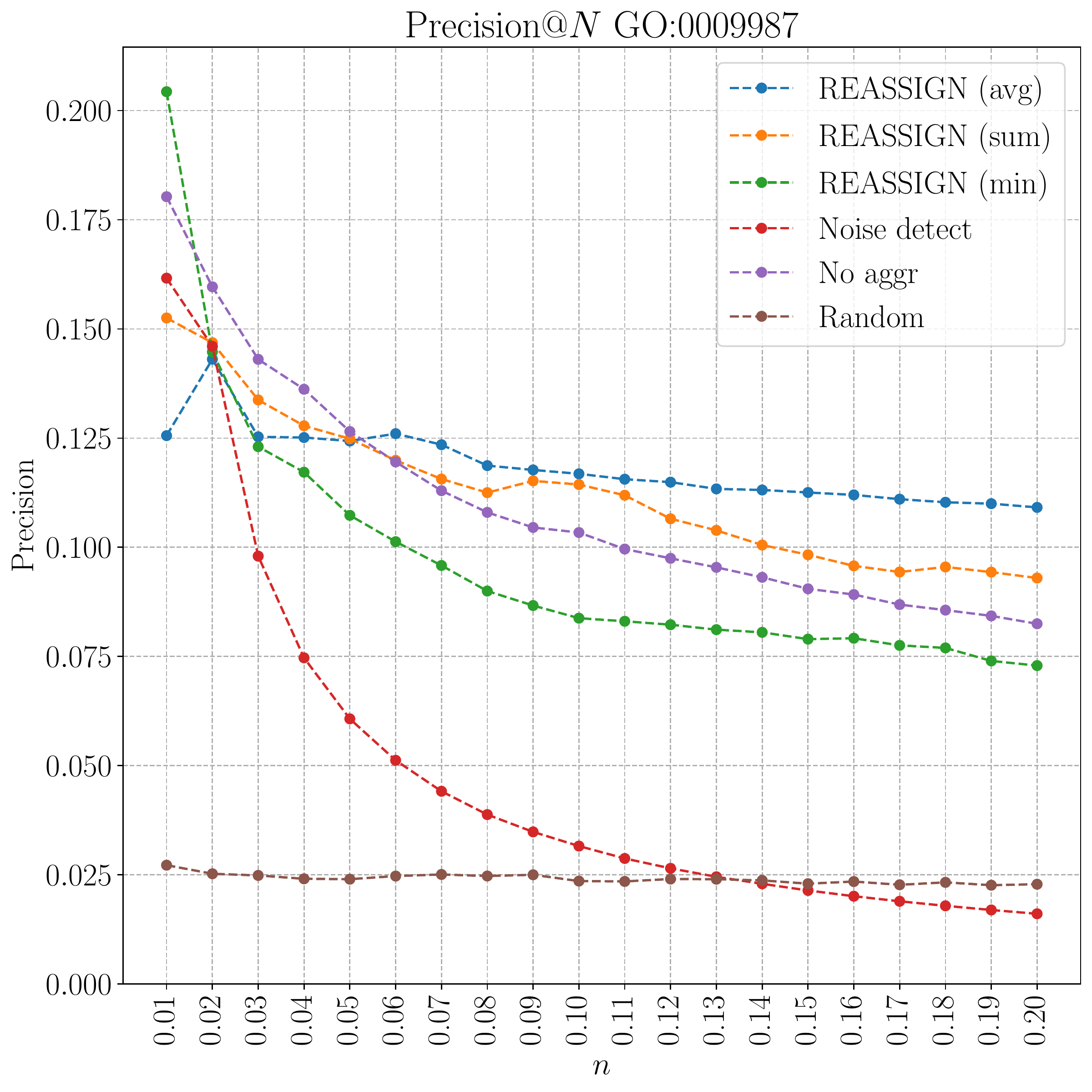}
	\caption{Precision@$N$ of all sub-hierarchies for 20 different
		values of $n$ ($N$ is derived from $n$) considering all evaluated
		methods. This Figure is best viewed in colors.}
	\label{fig:prec}
\end{figure*}

Figure~\ref{fig:prec} illustrates the predictive performance of all
sub-hierarchies measured with the precision@$N$. Sub-hierarchies are
shown in the same order of Table~\ref{tab:func}, from smallest (top
left) to largest (bottom right). The predictive performance of the
methods is measured based on a selection of the same number of
annotations for each dataset.

As can be seen, the variants of our method are mostly associated with
superior performance. More specifically, we highlight the results of
\textit{REASSIGN (avg)} on the datasets GO:0051179,
GO:0008152 and GO:0009987, where its curve is majoritarily above the
others considering most values of $n$. A noteworthy advantage in
performance is seen in the GO:0008152 dataset where \textit{REASSIGN
	(avg)} achieved 5\% higher precision than the closest competitor,
\textit{No aggr}, for all values of $n$.

The other variants of our method, \textit{REASSIGN (min)} and
\textit{REASSIGN (sum)}, also provided superior results in three
cases: GO:0006796, GO:0051179 and GO:0009987. Precisely, in the
GO:0006796 dataset, both methods are remarkably preferable over the
competitors due to superior performance in all values of $n$.
Likewise, these variants also yielded the best results in the
GO:0051179 (\textit{REASSIGN (sum)}) and GO:0009987 (\textit{REASSIGN
	(min)}) datasets when $n=0.01$ is considered. The performance of
these methods, associated with the performance of \textit{REASSIGN
	(avg)}, endorses the necessity of incorporating the hierarchical
relationships among the classes.

Complementary, \textit{No aggr}, the variant that overlooks the
hierarchy, managed to have the upperhand only in a few cases, such as
in the datasets GO:0032501, GO:0032502, GO:0019752 and GO:0065007,
specially when the value of $n$ is small. We suspect that this is
related to size of the sub-hierarchies. As presented in
Table~\ref{tab:func}, the datasets GO:0032501, GO:0032502 and
GO:0019752 have relatively smaller sub-hierarchies, thus incorporating
the hierarchy does not necessarily lead to better results. The
behaviour observed in GO:0065007 seems peculiar since it was the only
dataset where the method \textit{Noise detect} yields the highest
performance, followed by \textit{No aggr}, specially when smaller
values of $n$ are employed. However, it is worth mentioning that this
difference is marginal, as their performance ranges from approximately
6.5\% to 4\%, when compared to our proposed method using the average
as the aggregation method.

As opposed to that, \textit{Random} provides very underwhelming
results in most of the experiments where its performance is barely
superior than 0, making it negligible. Curiously, in some very
specific scenarios, \textit{Random} was capable of overcoming the
\textit{Noise detect} method, as seen in the GO:0032502, GO:0006796,
GO:0051179 and GO:0009987, when larger values of $n$ are analyzed. We
attribute this counter-intuitive finding to the performance of
\textit{Noise detect} as a whole. As shown in the
Figure~\ref{fig:prec}, there is a perceivable deterioration in
performance as the value of $n$ increases, whereas a less prominent
worsening was detected in the other methods.

Such deterioration is expected, since smaller values of
$n$ lead directly to smaller subsets of annotations to be evaluated,
artificially increasing the value of the precision. Hence,
selecting a smaller subset of annotations
does provide relatively better results.

Despite yielding superior results when compared to the literature,
our method shows that detecting missing annotations is a rather
challenging task, as seen in datasets such as GO:0065007 and
GO:0006796 where the best methods merely achieved 6\% and 8\% precision. We
suspect that the low availability of annotations, specially in
deeper levels of the hierarchy, plays a significant role in this
matter.

\subsection{Analysis of the area under the precision@$N$ curve}

Table~\ref{tab:auc} shows the area under the precision@$N$ curve for
all methods and sub-hierarchies. For each sub-hierarchy the best
method is highlighted with boldface. In all cases, variations of our
method always provide the highest AUP@NC. More specifically,
\textit{REASSIGN(avg)} provides the best performance in 4 datasets,
followed by \textit{No aggr} and \textit{REASSIGN(min)} on 3 and 1,
respectively. The competitor method \textit{Noise detect} did not
manage to have the best performance in any dataset.

Among the 3 cases where \textit{No aggr} had the upperhand, its
superiority was more pronounced only in one dataset, GO:0032501,
where it yielded 0.0208 AUP@NC and the second best method,
\textit{REASSIGN(avg)}, provided only 0.0137. In the other two cases,
GO:0032502 and GO:0065007, \textit{No aggr} was only marginally
better.

A different behaviour is seen in \textit{REASSIGN(avg)} where its
performance was considerably better in 3 out of the 4 cases.
Precisely, in GO:0051179, GO:0008152 and GO:0009987,
\textit{REASSIGN(avg)} provided visibly superior results in
comparison to the runner-up method. When compared solely against
\textit{Noise detect}, our most prominent variant,
\textit{REASSIGN(avg)} is consistently superior.

\begin{table*}[tbph!]
	\centering
	\resizebox{\textwidth}{!}{%
	\begin{tabular}{|l|r|r|r|r|r|r|}
		\hline
		Root &  \textit{REASSIGN (avg)} &  \textit{REASSIGN (sum)} &  \textit{REASSIGN (min)} &  \textit{No aggr} &  \textit{Random} &  \textit{Noise detect} \\\hline\hline
		GO:0032501 & 0.0137 & 0.0121 & 0.0114 & \textbf{0.0208} & 0.0063 & 0.0129\\
		GO:0019752 & \textbf{0.0141} & 0.0128 & 0.0113 & 0.0139 & 0.0027 & 0.0096\\
		GO:0032502 & 0.0111 & 0.0075 & 0.0111 & \textbf{0.0122} & 0.0033 & 0.0067\\
		GO:0006796 & 0.0040 & 0.0060 & \textbf{0.0071} & 0.0040 & 0.0028 & 0.0025\\
		GO:0051179 & \textbf{0.0286} & 0.0249 & 0.0139 & 0.0166 & 0.0047 & 0.0088\\
		GO:0065007 & 0.0057 & 0.0028 & 0.0055 & \textbf{0.0066} & 0.0022 & 0.0055\\
		GO:0008152 & \textbf{0.0339} & 0.0203 & 0.0191 & 0.0233 & 0.0048 & 0.0160\\
		GO:0009987 & \textbf{0.0225} & 0.0213 & 0.0180 & 0.0207 & 0.0046 & 0.0087\\\hline
	\end{tabular}
	}
	\caption{Area under the curve generated between the different values
		of $n$ (in the $x$-axis) and the precision@$N$ (in
		the $y$-axis), i.e., AUP@NC. The proposed method
		(\textit{REASSIGN (avg)}) outperforms \textit{No aggr} and
		\textit{Noise detect} methods in 5 and all sub-hierarchies,
		respectively.}
	\label{tab:auc}
\end{table*}

\begin{figure*}[tbph!]
	\centering
	\includegraphics[width=\linewidth]{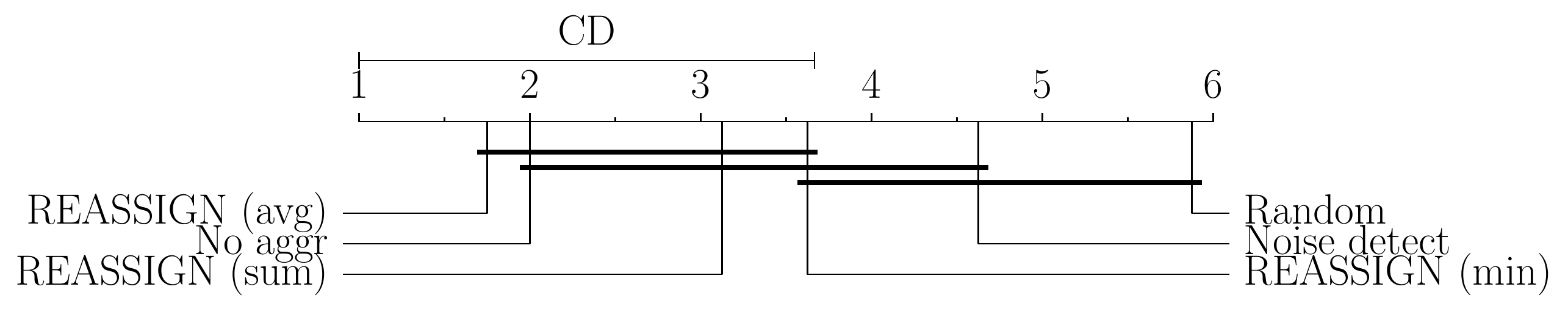}
	\caption{Friedman–Nemenyi test evaluating the area under the
		precision@$N$ curve, i.e., the curve generated between the different
		values of $n$ (in the $x$-axis) and the precision@$N$ (in the
		$y$-axis). Methods connected by a horizontal bar, of length less or
		equal to a critical distance, are not statistically significantly
		different. The proposed method is significantly different from the
		noise detection and the random methods.}
	\label{fig:nemenyi}
\end{figure*}

These results are further attested in the Friedman-Nemenyi presented
in Figure~\ref{fig:nemenyi}.  It can be seen that there is a
significant difference between \textit{REASSIGN (avg)} and the
competitor method \textit{Noise detect}, nonetheless no significant
difference is observed among the variant of our proposal.

Precisely, \textit{REASSIGN (avg)} is ranked in the first position
followed by \textit{No aggr}, \textit{REASSIGN (sum)} and
\textit{REASSIGN (min)}. The competitor method \textit{Noise
	detect}, the variant \textit{REASSIGN (min)} and Random are not
statistically significantly different.

\subsection{Comparison of true positives through the GO hierarchy levels}

To further evaluate the two best methods, \textit{REASSIGN (avg)} and
\textit{No aggr}, we have investigated their performance per level. More
specifically, we analyze the precision per level on two
datasets: GO:0032501 (Table~\ref{tab:prel32501}) and GO:0009987
(Table~\ref{tab:prel09987}). These were selected due to
their difference in hierarchy size (4 and 11 levels, respectively) and in performance.

\begin{table*}[tbph!]				
	\centering
	\begin{tabular}{|r|R{2cm}|R{4cm}|R{4cm}|}
		\hline
		Level & \# $0\rightarrow 1$ & 
		\textit{REASSIGN (avg)} & \textit{No aggr} \\ 
		\hline\hline
		1     & 97  & 10/77 (\textbf{12.99\%}) & 26/185 (12.05\%) \\
		2     & 51  & 6/87 ( 6.90\%)    & 3/24 (\textbf{12.50\%})    \\
		3     & 33  & 1/50 (\textbf{0.02\%})   & 0/5 ( 0\%)        \\
		4     & 3   & 0/0 ( 0\%)       & 0/0 ( 0\%)        \\\hline\hline
		Total & 184 & 17/214 ( 7.94\%) & \textbf{29/214 (13.55\%)} \\\hline
		
	\end{tabular}
	\caption{Number of predicted annotations per level for the proposed
		\textit{REASSIGN (avg)} and \textit{No aggr} methods for the
		sub-hierarchy GO:0032501. The second column shows the number of
		missing annotations (0s that became 1s) per level, followed by the
		number of true positives, the number of predicted annotations and
		the precision per level for both methods. The last row shows the
		total number of missing annotations in the sub-hierarchy and the
		total number of true positives predicted by each method.}
	\label{tab:prel32501}
	\bigskip
	\begin{tabular}{|r|R{2cm}|R{4cm}|R{4cm}|}
		\hline
		Level & \# $0\rightarrow 1$ & 
		\textit{REASSIGN (avg)} & \textit{No aggr} \\ 
		\hline\hline
		1     & 2334 & 581/2871 (\textbf{20.24\%}) & 261/1913 (13.64\%) \\
		2     & 3976 & 502/3121 (\textbf{16.08\%}) & 938/7458 (12.58\%) \\
		3     & 3833 & 302/2748 (\textbf{10.99\%}) & 221/2025 (10.91\%) \\
		4     & 2402 & 202/2197 (\textbf{9.19\%})  & 116/1268 ( 9.15\%)  \\
		5     & 1729 & 87/1625 ( 5.35\%)   & 76/793 (\textbf{ 9.58\%)}    \\
		6     & 1109 & 20/699 ( 2.86\%)    & 6/52 (\textbf{11.54\%})     \\
		7     & 721  & 12/271 ( 4.43\%)    & 5/46 (\textbf{10.87\%})     \\
		8     & 304  & 6/34 (\textbf{17.65\%})     & 1/16 ( 6.25\%)      \\
		9     & 85   & 0/19 (  0\%)         & 0/14 ( 0\%)         \\
		10    & 24   & 0/0 (  0\%)          & 0/0 ( 0\%)          \\
		11    & 3    & 0/0 (  0\%)          & 0/0 ( 0\%)          \\\hline\hline
		Total & \numprint{16520} & \textbf{1712/\numprint{13585} (12.60\%)} & 1624/\numprint{13585} (11.95\%) \\\hline
	\end{tabular}
	\caption{Number of predicted annotations per level for the proposed
		\textit{REASSIGN (avg)} and \textit{No aggr} methods for the
		sub-hierarchy GO:0009987. The second column shows the number of
		missing annotations (0s that became 1s) per level, followed by the
		number of true positives, the number of predicted annotations and
		the precision per level for both methods. The last row shows the
		total number of missing annotations in the sub-hierarchy and the
		total number of true positives predicted by each method.}
	\label{tab:prel09987}		
\end{table*}

As can be seen in Table~\ref{tab:prel32501}, \textit{No aggr} focuses
substantially on annotations present in the first level of the
hierarchy where it correctly predicts 26 annotations, whereas
\textit{REASSIGN (avg)} managed to obtain only 10. In the second and
third level, however, \textit{REASSIGN (avg)} was capable of
accurately identifying more missing annotations. We believe that the
aggregation function is responsible for this difference, as classes
located in deeper levels are often associated to very low prediction
probabilities, making their selection very unlikely by \textit{No aggr}. 

A slightly different behaviour in performance is noticed at
Table~\ref{tab:prel09987} where \textit{REASSIGN (avg)} had the upper
hand with 1712 over 1624 provided by \textit{No aggr}. Despite of
that, a similar tendency in the distribution of the annotations was
noticed: the missing annotations identified by \textit{No aggr} are
mostly located in the shallow levels of the sub-hierarchy, specially on
the second level in this case, whereas \textit{REASSIGN (avg)} seeks
deeper annotations. Nevertheless, \textit{REASSIGN (avg)} detects in
average the double of missing annotations than \textit{No aggr} in all
levels, except for the second one.

Hence, we may assume that \textit{No aggr} is more likely to provide
desirable results when sub-hierarchies with fewer levels (in this case,
4) are considered. As opposed to that, employing the average as the
aggregation function is preferred when deeper, and possibly more
complex, sub-hierarchies are investigated. However, it is worth
mentioning that detecting missing annotations in deeper levels still
remains a challenge since no method was able to detect them in the
deepest level of both hierarchies.

\section{Conclusion and Future Work}
\label{sec:concl}

In this work, we have presented a novel method to detect missing
annotations in HMC datasets. More specifically, we proposed a method,
with 3 possible variants, that exploits the class hierarchy by
computing aggregated probabilities (e.g., average, sum and minimum)
of the paths of classes from the leaves to the root for each
instance. Furthermore, the proposed method is presented in the context of
predicting missing gene function annotations, where these aggregated
probabilities are further used to select a set of annotations to be
verified through \textit{in vivo} experiments.

Experiments on \textit{Oriza sativa Japonica}, a variety of rice,
showcased that our proposed method yields superior results when
compared to competitor methods from the literature. Furthermore, we
could also identify that incorporating the hierarchy of classes into
the method often improves the results. Precisely, averaging the
probabilities leads to the identification of missing annotations in
deeper levels of the hierarchy, which is often regarded as more
informative. Despite of that, our results also highlight how
challenging this task is.

Hence, as future work, we highlight two main lines. First,
considering other ways of aggregating the probabilities, may lead to
improved results, specially on deeper levels of the hierarchy.
Additionally, we may also explore transfer learning, specifically
domain adaptation, to enrich the training of HMC classifiers. Using
information from other organisms can not only result in higher
prediction performance, but may also allow the application of HMC to
organisms without functional information available.

Even though the proposed method is focused in detecting missing
annotations in the datasets, detecting annotations that were removed
instead of added, may be of interest to identify wrong associations
and to improve the quality of the datasets.

\section*{Acknowledgements}
This work was funded by the OMICAS program
anchored at the Pontificia
Universidad Javeriana in Cali and funded within the Colombian
Scientific Ecosystem by The World Bank, the Colombian Ministry of
Science, Technology and Innovation, the Colombian Ministry of
Education and the Colombian Ministry of Industry and Turism, and
ICETEX, under GRANT ID: FP44842-217-2018. The authors also
acknowledge the support from the Research Fund Flanders (through
research project G080118N) and from the Flemish Government (AI
Research Program).

\bibliographystyle{cas-model2-names}
\bibliography{main}

\end{document}